\documentclass[letterpaper]{article} 
\usepackage{aaai25}  
\usepackage{times}  
\usepackage{helvet}  
\usepackage{courier}  
\usepackage[hyphens]{url}  
\usepackage{graphicx} 
\usepackage{natbib}  
\usepackage{caption} 
\usepackage{algorithm}
\usepackage{algorithmic}

\usepackage{newfloat}
\usepackage{listings}

\usepackage{amsmath,amsfonts,bm,amssymb,multirow,booktabs}
\usepackage{pifont} 
\usepackage{tablefootnote}
\DeclareCaptionStyle{ruled}{labelfont=normalfont,labelsep=colon,strut=off} 
\floatstyle{ruled}
\newfloat{listing}{tb}{lst}{}
\floatname{listing}{Listing}
\title{What Makes In-context Learning Effective for Mathematical \\Reasoning: A Theoretical Analysis}
\author{
    Jiayu Liu\textsuperscript{\rm 1,2}, 
    Zhenya Huang\textsuperscript{\rm 1,2},
    Chaokun Wang\textsuperscript{\rm 1,2},
    Xunpeng Huang\textsuperscript{\rm 3},\\
    Chengxiang Zhai\textsuperscript{\rm 4},
    Enhong Chen\textsuperscript{\rm 1,2}
}
\affiliations{
    \textsuperscript{\rm 1}University of Science and Technology of China\\
    \textsuperscript{\rm 2}State Key Laboratory of Cognitive Intelligence\\
    \textsuperscript{\rm 3}Hong Kong University of Science and Technology\\
    \textsuperscript{\rm 4}University of Illinois at Urbana-Champaign

}

\usepackage{bibentry}
\begin{document}

\maketitle

\begin{abstract}
Owing to the capability of in-context learning, large language models (LLMs) have shown impressive performance across diverse mathematical reasoning benchmarks. However, we find that few-shot demonstrations can sometimes bring negative performance and their effectiveness on LLMs' reasoning abilities remains unreliable. To this end, in this paper, we aim to theoretically analyze the impact of in-context demonstrations on LLMs' reasoning performance. We prove that the reasoning efficacy (measured by empirical prediction loss) can be bounded by a \emph{LLM-oriented semantic similarity} and an \emph{inference stability of demonstrations}, which is general for both one-shot and few-shot scenarios. Based on this finding, we propose a straightforward, generalizable, and low-complexity demonstration selection method named LMS3. It can adaptively facilitate to select the most pertinent samples for different LLMs and includes a novel demonstration rejection mechanism to automatically filter out samples that are unsuitable for few-shot learning. Through experiments on three representative benchmarks, two LLM backbones, and multiple few-shot settings, we verify that our LMS3 has superiority and achieves consistent improvements on all datasets, which existing methods have been unable to accomplish.
\end{abstract}

%
\section{Introduction}
Mathematical reasoning is a critical task and serves as a milestone in assessing the progress of artificial intelligence~\cite{zhang2020gap,liu2023learning}. Currently, many large language models (LLMs) have exhibited strong performance across various mathematical reasoning benchmarks~\cite{hendrycks2021measuring,cobbe2021training}. A key capability of these LLMs is in-context learning (ICL)~\cite{dong2022survey}, which enables them to learn from a few examples to implement specific logical structures~\cite{wei2022chain} or utilize codes~\cite{chen2023program} to improve reasoning accuracy. Based on this ability, they can adeptly address a wide variety of problems across different types and difficulty, ranging from elementary word problems to college-level algebra~\cite{brown2020language,achiam2023gpt}.

\begin{figure}[t]
\centering
\setlength{\abovecaptionskip}{2pt}
\includegraphics[width=0.49\linewidth]{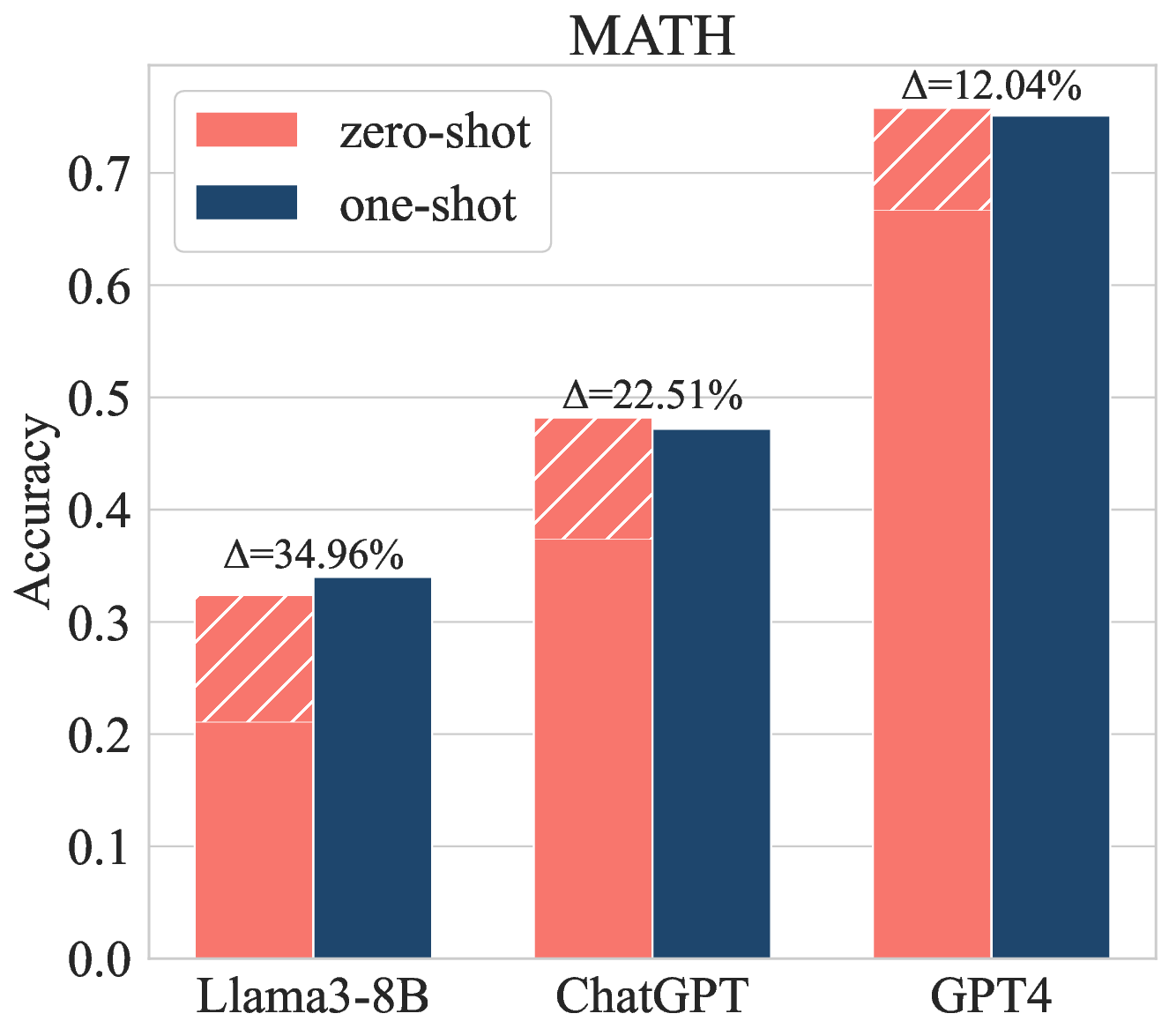}
\includegraphics[width=0.49\linewidth]{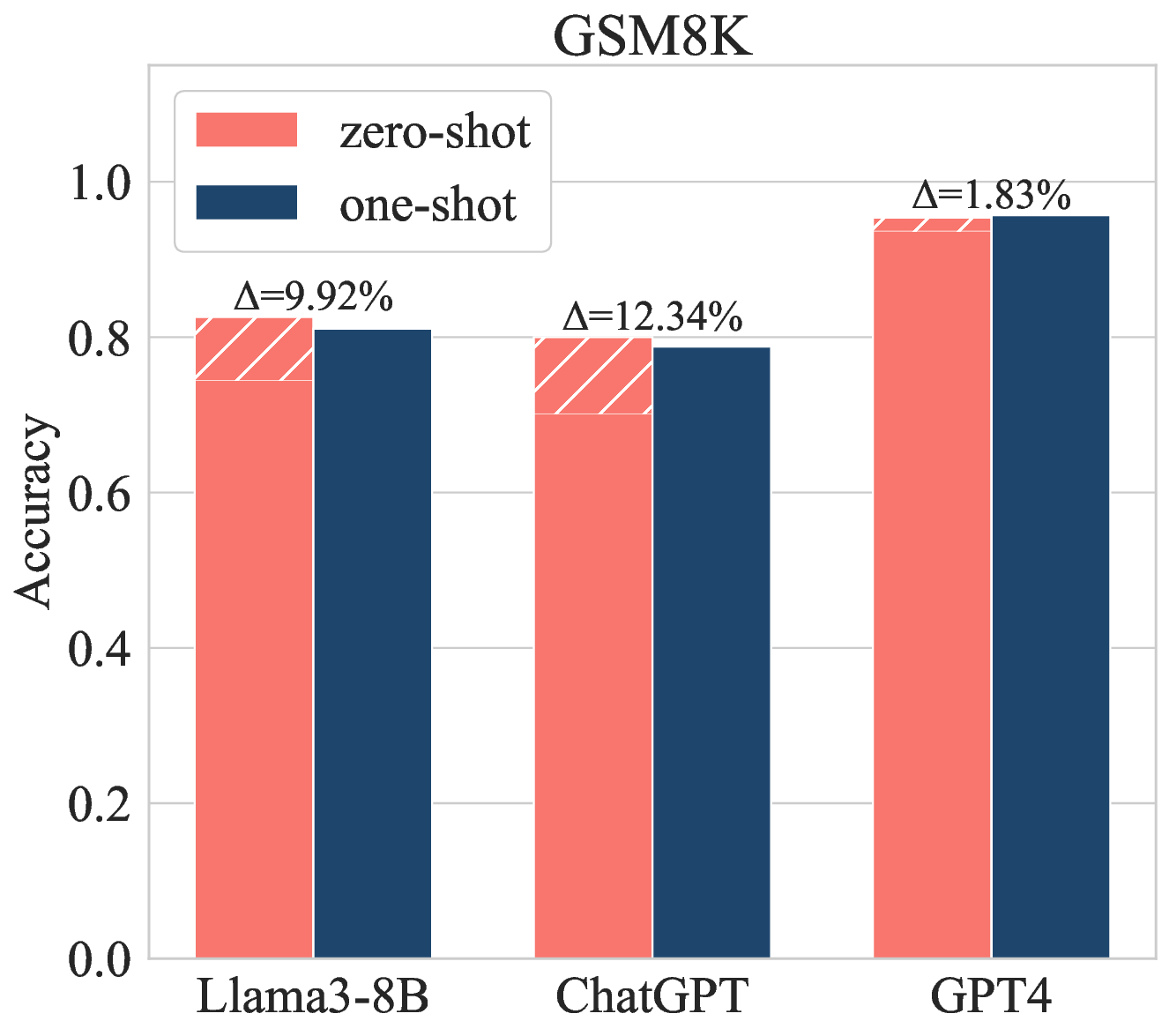}
\caption{Problem-solving Accuracy of zero-shot and one-shot settings. The hatched areas represent that in the one-shot setting, the model answers incorrectly $\Delta$ proportion of problems that are answered correctly in the zero-shot setting.}
\label{pre_exp}
\vspace{-15pt}
\end{figure}

However, it remains an unresolved issue whether in-context learning truly enhances LLMs' mathematical reasoning abilities. To show this phenomenon, in Figure~\ref{pre_exp}, we present the zero-shot and one-shot results of Llama3-8B~\cite{meta2024introducing}, ChatGPT, and GPT-4~\cite{achiam2023gpt} on two representative benchmarks MATH~\cite{hendrycks2021measuring} and GSM8K~\cite{cobbe2021training}. Surprisingly, we find that 1) When given an example, their problem-solving accuracy does not consistently improve, and sometimes even declines (e.g., ChatGPT on MATH dataset). 2) Further analysis reveals that in the one-shot setting, these LLMs even fail in 1.83\%-34.96\% of problems (marked with white hatching) that they have correctly solved in the zero-shot scenario. This raises an important research question: \emph{Is in-context learning always effective for mathematical reasoning, and under what conditions does it work?}

To address this issue, existing literature primarily analyzes the impact of demonstrations from an empirical perspective. For instance, researchers have revealed several important factors, including the similarity to test samples~\cite{liu2022makes}, the diversity~\cite{gao2024customizing}, complexity~\cite{an2023context} and perplexity~\cite{sorensen2022information} of demonstrations, as well as the Inference LLM used~\cite{peng2024revisiting,ye2023compositional}. Regarding the theoretical foundations of ICL, existing work has tried to explain the introduction of demonstrations as a form of meta-gradient optimization~\cite{dai2023can}, kernel regression~\cite{han2023explaining}, and token reinforcement~\cite{yanunderstanding}. However, these studies 1) have not provided precise quantification of the impact of demonstrations on LLMs' reasoning performance, nor 2) have they offered theoretical conclusions on when demonstrations are beneficial.

To this end, in this paper, we first theoretically analyze the impact of a demonstration on the reasoning performance in one-shot scenario. Our theory indicates that a sufficient condition for one-shot to outperform zero-shot is that 1) \emph{the demonstration and test sample have similar semantics encoded by the inference LLM} and that 2) \emph{the inference LLM exhibits sufficient stability in reasoning the answer of the demonstration itself}. The former goes beyond traditional methods that rely solely on the semantic similarity between demonstrations and test samples, pointing out the critical role of the inference LLM's encoding capacity in its parameters, while also being generalizable to these methods. The latter introduces the concept and measurement of \emph{inference stability of demonstrations} for the first time. It should be emphasized that our theory is general and we further extend it to the $k$-shot scenario. 

Based on our theory, we propose a simple yet effective demonstration selection method, named \textbf{LMS3}, to balance the \emph{\textbf{L}L\textbf{M}-oriented \textbf{S}emantic \textbf{S}imilarity} and \emph{inference \textbf{S}tability of demonstrations}, allowing for the automatic selection of approximately optimal samples tailored to different LLMs. Additionally, to ensure that the sufficient condition of our theories is essentially satisfied, we introduce an innovative demonstration rejection mechanism that can adaptively identify when few-shot learning should \emph{not} be used, which is the first attempt in the field. Our method has strong theoretical advantages, generalization ability, and low complexity. Experiments on three benchmarks demonstrate its consistent improvements in both one-shot and few-shot scenarios. The contributions of this paper are as follows:
\begin{itemize}
  \item We theoretically quantify the effect of demonstrations on ICL reasoning performance under one/few-shot settings. We prove that it can be bounded by \emph{LLM-oriented semantic similarity} and \emph{inference stability of demonstrations}.
  \item We propose a novel demonstration selection method, LMS3, which can generalize to various existing methods and offers better scalability and complexity.
  \item We validate our method on three mathematical benchmarks using multiple LLMs as backbones, demonstrating improvements in problem-solving accuracy, generalization ability, and interpretability.
\end{itemize}
\section{Related Work}
\textbf{Mathematical Reasoning.} Mathematical reasoning is a critical benchmark for assessing the level of artificial intelligence~\cite{zhang2020gap,liu2023learning}. Early work in this area mainly focused on rule-based, template-based, and statistical machine learning methods for simple math word problems~\cite{feigenbaum1963computers,fletcher1985understanding}. With the development of large language models (LLMs), current mathematical reasoning work primarily falls into two categories. The first category improves the mathematical reasoning capabilities of general LLMs through techniques such as prompt engineering. On one hand, they endow LLMs with chain-like~\cite{kojima2022large}, tree-like~\cite{yao2024tree}, or graph-like~\cite{besta2024graph} reasoning processes, or require LLMs to generate code~\cite{chen2023program,gao2023pal} to address potential numerical computation errors. On the other hand, they also involve providing the model with certain examples in the prompts through retrieval-augmented generation~\cite{wei2022chain,asaiself}, allowing the model to solve problems based on similar approaches using its contextual learning abilities. The second category is to fine-tune a specific mathematical LLM using mathematical problem-solving data~\cite{lewkowycz2022solving,yuemammoth}. This type of work addresses both the diversity of mathematical problems (e.g., range from elementary to university-level difficulties~\cite{hendrycks2021measuring}, cover various types~\cite{trinh2024solving}, rephrase original corpus~\cite{yumetamath}) and the problem-solving process itself (e.g., supervise the training with the reasoning steps, rather than relying solely on the final answers~\cite{lightman2023let,luo2023wizardmath}). 

\textbf{In-context Learning.} In-context Learning (ICL) focuses on making LLMs learn and reason based on existing examples~\cite{dong2022survey}. Its advantage lies in the adaptability and flexibility for different tasks and scenarios. However, the selection of examples remain a central challenge, where current researches have developed supervised methods and unsupervised methods. This paper focuses on unsupervised methods, which can be grouped into three main categories. The first and currently most prominent method is called Similar-ICL~\cite{liu2022makes,luo2023dr,zhang2023retrieve,fu2022complexity}, which aims to find examples with closest semantic representations to the test sample. The semantic representation approaches include TF-IDF, BM25~\cite{robertson2009probabilistic}, T5 encoding~\cite{raffel2020exploring}, BGE-M3~\cite{chen2024bge}, OpenAI embedding, etc. The second line of methods calculate the impact of each demonstration on the test sample~\cite{peng2024revisiting}. Impact calculation approaches include influence function~\cite{van2024context,chang2023data}, mutual information~\cite{sorensen2022information}, perplexity~\cite{gonen2023demystifying}, code-length~\cite{wu2023self}, etc. The third category uses the feedback from LLMs to dynamically select demonstrations~\cite{nguyen2023context,qin2023context}. Regarding the underlying mechanisms of ICL, most existing research explored the impact of empirical factors such as the number of examples, gold labels, diversity, and types of LLMs from an experimental perspective~\cite{pan2023context,peng2024revisiting,min2022rethinking}. Some theoretical explorations explain ICL from perspectives including meta-gradient updates~\cite{dai2023can}, kernel regression~\cite{han2023explaining}, and token reinforcement~\cite{yanunderstanding}. In comparison, to the best of our knowledge, we are the first to theoretically quantify the impact of demonstrations on reasoning performance and identify when they are effective.
%
\section{Theoretical Analysis}\label{section_theory}
\paragraph{Notations.}In in-context learning (ICL) setup, we have a demonstration pool $\mathcal{D}$ and a test set $\mathcal{D}_{test}$, which contain $\mathcal{M}$ and $\mathcal{N}$ mathematical problems respectively. The $k$-shot in-context learning is formulated as appending $k$ demonstrations $\{(X_1,y_1),(X_2,y_2),...,(X_k,y_k)\}\subseteq \mathcal{D}$ with the test data $X_{test}\in \mathcal{D}_{test}$ in prompt to reason the solution 
\begin{small}
\begin{equation}
\hat{y}_{test}\stackrel{def}{=}LLM((X_1,y_1),(X_2,y_2),...,(X_k,y_k), X_{test}),
\end{equation}
\end{small}
\noindent where $X_i, X_{test}$ represent the problem context and $y_i$ represents the labeled solution. The prediction loss on $X_{test}$ is denoted as $L(X_{test}, y_{test})$. In the following, we omit the symbol $y$ and use $X$ to express each demonstration for brevity.

To evaluate the influence of a demonstration $X$ on inferencing the answer of $X_{test}$, we use $\mathbf{h},\ \mathbf{h}_{test}\in \mathbb{R}^d$ to denote the representation of problem $X$ and $X_{test}$. Then, the Transformer attention mechanism in ICL setting is denoted as:
\begin{small}
\begin{equation}
\begin{aligned}\label{attention_icl}
  &\mathcal{F}_{ICL}(\mathbf{h}_{test})=Attn(V,K,Q,\mathbf{h}_{test}) \\
  &=W_V[\mathbf{h},\mathbf{h}_{test}]\cdot softmax\left(\frac{(W_K[\mathbf{h},\mathbf{h}_{test}])^{\mathrm{T}} \cdot W_Q\mathbf{h}_{test}}{\sqrt{d}}\right),
\end{aligned}
\end{equation}
\end{small}
where $W_Q,W_K,W_V$ are the projection matrices for computing the attention queries, keys, and values, respectively. Without loss of generality, we omit $W_Q$ in $\mathcal{F}_{ICL}(\mathbf{h}_{test})$ because we can redefine $W_K=W_K^{\mathrm{T}}\cdot W_Q$. As a result, we only keep $W_K\in \mathbb{R}^{d\times d},W_V\in \mathbb{R}^{d^\prime\times d}$ in our setting, where $d^\prime$ is the output dimension of layer $\mathcal{F}_{ICL}$. Following Dai et al.~\shortcite{dai2023can}, we approximate the attention to linear attention by removing the softmax function:
\begin{small}
\begin{equation}
\begin{aligned}\label{attention_icl_appro}
  &\mathcal{F}_{ICL}(\mathbf{h}_{test})\approx W_V[\mathbf{h},\mathbf{h}_{test}]\cdot \left(\frac{(W_K[\mathbf{h},\mathbf{h}_{test}])^{\mathrm{T}} \cdot \mathbf{h}_{test}}{\sqrt{d}}\right) \\
  &=\frac{W_V}{\sqrt{d}}\mathbf{h}_{test}\cdot (W_K\mathbf{h}_{test})^{\mathrm{T}}\cdot \mathbf{h}_{test} + \frac{W_V}{\sqrt{d}}\mathbf{h}\cdot (W_K\mathbf{h})^{\mathrm{T}}\cdot \mathbf{h}_{test}.\\
\end{aligned}
\end{equation}
\end{small}

\paragraph{Analogy to Linear Optimization.} We start our analysis of Eq.~\eqref{attention_icl_appro} by considering a linear function $\mathcal{F}(z)\stackrel{def}{=} W\cdot z, W\in \mathbb{R}^{d^\prime\times d}, z\in \mathbb{R}^{d}$. Specifically, given $\mathcal{F}(z)$ with an initialized parameters $W_0$, assume we have a training data $z_0\in \mathbb{R}^{d}$, then the gradient of loss $L(\mathcal{F})$ can be written as $\Delta W=\nabla_\mathcal{F} L(z_0,W_0)\cdot z_0^{\mathrm{T}}$. Applying the gradient to parameter optimization, the prediction of a test sample $\mathbf{h}_{test}$ is $\mathcal{F}(\mathbf{h}_{test})=W_0\cdot \mathbf{h}_{test} + \nabla_\mathcal{F} L(z_0,W_0)\cdot z_0^{\mathrm{T}}\cdot\mathbf{h}_{test}$. 

Based on this idea, Eq.~\eqref{attention_icl_appro} can be interpreted as: 1) We have a linear function $\mathcal{F}(z)$ with initialized parameters 
\begin{small}
\begin{equation}\label{init_para}
W_0=\frac{W_V}{\sqrt{d}}\mathbf{h}_{test}\cdot (W_K\mathbf{h}_{test})^{\mathrm{T}}.
\end{equation}
\end{small}
2) We introduce a training data $z_0=W_K\mathbf{h}$ to optimize the parameters, with the gradient at $(z_0,W_0)$ satisfies:
\begin{small}
\begin{equation}\label{meta_gradient}
\nabla_\mathcal{F} L(z_0,W_0)= \frac{W_V}{\sqrt{d}}\mathbf{h}.
\end{equation}
\end{small}
3) We finally apply the optimized parameters to calculate the result of test data $\mathbf{h}_{test}\in \mathcal{D}_{test}$. 

Under this setting, we aim to estimate the influence of the data $z_0=W_K\mathbf{h}$ (corresponds to demonstration $X\in \mathcal{D}$) on loss $L(\mathcal{F}(\mathbf{h}_{test}))$. Before detailed derivation, we first give three mathematical annotations:
\begin{small}
\begin{equation}\label{annotation}
\begin{gathered}
  \hat{W}\stackrel{def}{=}\operatorname*{argmin}_W \frac{1}{|\mathcal{D}_{pre}|}\Sigma_{z\in \mathcal{D}_{pre}}L(\mathcal{F}(z))\\
  \hat{W}_{\epsilon,z_0}\stackrel{def}{=}\operatorname*{argmin}_W \frac{1}{|\mathcal{D}_{pre}|}\Sigma_{z\in \mathcal{D}_{pre}}L(\mathcal{F}(z))+\epsilon\cdot L(\mathcal{F}(z_0))\\
  H_{\hat{W}}=\frac{1}{|\mathcal{D}_{pre}|}\Sigma_{z\in \mathcal{D}_{pre}}\nabla^2_WL(z,\hat{W}),
\end{gathered}
\end{equation}
\end{small}
where $\mathcal{D}_{pre}$ is the data for pre-training a LLM, and $H_{\hat{W}}$ is the Hessian matrix which is positive definite by assumption~\cite{van2024context}. It is worth noting that the pre-trained parameters $\hat{W}$ are actually the initialized parameters in our above setting, i.e., $\hat{W}=W_0$. Taking $\epsilon=\frac{1}{|\mathcal{D}_{pre}|}$, the testing loss on $\mathbf{h}_{test}$ is represented as $L(\mathbf{h}_{test},\hat{W}_{\frac{1}{|\mathcal{D}_{pre}|},z_0})$. On this basis, we derive the following theorem:

\textbf{Theorem 1.} Assume $\nabla_\mathcal{F}L$ is Lipschitz continuous w.r.t $\mathcal{F}$ with constant $\mu$. If inequality~\eqref{condition} holds true, then $L(\mathbf{h}_{test},\hat{W}_{\frac{1}{|\mathcal{D}_{pre}|},z_0})<L(\mathbf{h}_{test},\hat{W}_{0,z_0})$, i.e., introducing the training sample $z_0$ (i.e., demonstration $X$) can reduce the testing loss on $\mathbf{h}_{test}$. $\frac{1}{\lambda_{dd^\prime}}, \frac{1}{\lambda_1}$ are the largest and smallest eigenvalues of $H_{\hat{W}}$, respectively.
\begin{small}
\begin{equation}\label{condition}
\begin{gathered}
\frac{\lambda_{dd^\prime}}{\lambda_1}\|\nabla_W L(\mathbf{h}_{test},\hat{W})\|> \|\mathbf{h}_{test}-z_0\|\cdot(\|\frac{W_V}{\sqrt{d}}\mathbf{h}\|+\mu C_1)\\
C_1=\|\frac{W_V}{\sqrt{d}}\mathbf{h}_{test}\|\cdot \|W_K\mathbf{h}_{test}\|\cdot\|\mathbf{h}_{test}\|
\end{gathered}
\end{equation}
\end{small}

We refer the readers to \emph{Appendix} 1 for the detailed proof and explanations, and present the sketch here.

\emph{Proof}. With $\hat{W},\hat{W}_{\epsilon,z_0}$, the influence of upweighting $z_0$ on the empirical loss is~\cite{ling1984residuals,koh2017understanding}:
\begin{small}
\begin{equation}\label{influence_loss}
\begin{aligned}
 \mathcal{I}_{loss}(z)&=\left.\frac{dL(\mathbf{h}_{test},\hat{W}_{\epsilon,z_0})}{d\epsilon}\right|_{\epsilon=0}\\
 &=-\nabla_W L(\mathbf{h}_{test},\hat{W})^{\mathrm{T}}\cdot H^{-1}_{\hat{W}}\nabla_W L(z_0,\hat{W}).
\end{aligned}
\end{equation}
\end{small}
Then, the testing loss $L(\mathbf{h}_{test},\hat{W}_{\frac{1}{|\mathcal{D}_{pre}|},z_0})$ can be evaluated by Taylor approximation since $\frac{1}{|\mathcal{D}_{pre}|}$ is sufficiently small:
\begin{small}
\begin{equation}\label{taylor}
\begin{aligned}
  &L(\mathbf{h}_{test},\hat{W}_{\frac{1}{|\mathcal{D}_{pre}|},z_0})\approx L(\mathbf{h}_{test},\hat{W}_{0,z_0})+\frac{1}{|\mathcal{D}_{pre}|}\left.\frac{dL(\mathbf{h}_{test},\hat{W}_{\epsilon,z_0})}{d\epsilon}\right|_{\epsilon=0}\\
  &=L(\mathbf{h}_{test},\hat{W}_{0,z_0})-\frac{1}{|\mathcal{D}_{pre}|}\nabla_W L(\mathbf{h}_{test},\hat{W})^{\mathrm{T}}\cdot H^{-1}_{\hat{W}}\nabla_W L(z_0,\hat{W}).
\end{aligned}
\end{equation}
\end{small}

Therefore, now the question turns to evaluate 
\begin{small}
\begin{equation}\label{L_1}
\begin{aligned}
L_1&\stackrel{def}{=}\nabla_W L(\mathbf{h}_{test},\hat{W})^{\mathrm{T}}\cdot H^{-1}_{\hat{W}}\nabla_W L(z_0,\hat{W})\\
=&\underbrace{(\nabla_W L(z_0,\hat{W})-\nabla_W L(\mathbf{h}_{test},\hat{W}))^{\mathrm{T}}\cdot H^{-1}_{\hat{W}}\nabla_W L(\mathbf{h}_{test},\hat{W})}_{L_{11}}\\
  &+\underbrace{\nabla_W L(\mathbf{h}_{test},\hat{W})^{\mathrm{T}}\cdot H^{-1}_{\hat{W}}\nabla_W L(\mathbf{h}_{test},\hat{W})}_{L_{12}}
\end{aligned}
\end{equation}
\end{small}

Since $H_{\hat{W}}$ is positive definite, we denote $\lambda_1\geq\lambda_2\geq...\geq\lambda_{dd^\prime}>0$ are the eigenvalues of $H^{-1}_{\hat{W}}$ and can prove that
\begin{small}
\begin{equation}\label{l11}
\begin{aligned}
  L_{11} &\geq -\lambda_1\|\nabla_W L(\mathbf{h}_{test},\hat{W})\|\cdot\left(\|\nabla_\mathcal{F} L(\mathbf{h}_{test},\hat{W})-\nabla_\mathcal{F} L(z_0,\hat{W})\|\right.\\
  &\quad\cdot \left.\|\mathbf{h}_{test}\|+\|\nabla_\mathcal{F} L(z_0,\hat{W})\|\cdot \|\mathbf{h}_{test}-z_0\|\right),\\ 
\end{aligned}
\end{equation}
\end{small}
Since $\nabla_\mathcal{F}L$ is Lipschitz continuous, we get $L_{11}\geq$
\begin{small}
\begin{equation}\label{l11_1}
\begin{aligned}
&-\lambda_1\|\nabla_W L(\mathbf{h}_{test},\hat{W})\|\cdot (\mu \|\hat{W}(\mathbf{h}_{test}-z_0)\|\cdot\|\mathbf{h}_{test}\|+\\
&\|\nabla_\mathcal{F} L(z_0,\hat{W})\|\cdot \|\mathbf{h}_{test}-z_0\|)
\end{aligned}
\end{equation}
\end{small}

Applying Eqs.~\eqref{init_para} and~\eqref{meta_gradient} into Eq.~\eqref{l11_1}, we have:
\begin{small}
\begin{equation}\label{l11_2}
\|\hat{W}(\mathbf{h}_{test}-z_0)\|\leq\|\frac{W_V}{\sqrt{d}}\mathbf{h}_{test}\|\cdot \|W_K\mathbf{h}_{test}\|\cdot\|\mathbf{h}_{test}-z_0\|
\end{equation}
\end{small}
\begin{small}
\begin{equation}\label{l11_3}
\|\nabla_\mathcal{F} L(z_0,\hat{W})\|\cdot \|\mathbf{h}_{test}-z_0\|=\|\frac{W_V}{\sqrt{d}}\mathbf{h}\|\cdot \|\mathbf{h}_{test}-z_0\|
\end{equation}
\end{small}

For $L_{12}$, we similarly prove that:
\begin{small}
\begin{equation}\label{l12}
 L_{12}=\Sigma_{i=1}^{dd^\prime}\lambda_ib^2_i\geq \lambda_{dd^\prime}\|\nabla_W L(\mathbf{h}_{test},\hat{W})\|^2
\end{equation}
\end{small}

Combining Eqs.~\eqref{l11_1}-\eqref{l12}, we finally get:
\begin{small}
\begin{equation}\label{L1_final1}
\begin{aligned}
L_1 \geq&\lambda_{dd^\prime}\|\nabla_W L(\mathbf{h}_{test},\hat{W})\|^2-\lambda_1\|\nabla_W L(\mathbf{h}_{test},\hat{W})\|\cdot\left(\mu \cdot C_1\right.\\
  & \cdot \|\mathbf{h}_{test}-z_0\|\cdot\|\mathbf{h}_{test}\|+\|\frac{W_V}{\sqrt{d}}\mathbf{h}\|\cdot \left.\|\mathbf{h}_{test}-z_0\|\right).
\end{aligned}
\end{equation}
\end{small}

According to Eq.~\eqref{condition}, the right-hand side of Eq.~\eqref{L1_final1} is greater than $0$, which leads to the conclusion. $\Box$

\paragraph{Extension to \emph{k}-shot setting.} In Theorem 1, we only consider one demonstration $X$ (i.e., the one-shot scenario). For the $k$-shot scenario, Eq~\eqref{attention_icl_appro} can be written as
\begin{small}
\begin{equation}
\begin{aligned}\label{attention_icl_appro_k}
  \mathcal{F}^k_{ICL}(\mathbf{h}_{test})\approx &\frac{W_V}{\sqrt{d}}\mathbf{h}_{test}\cdot (W_K\mathbf{h}_{test})^{\mathrm{T}}\cdot \mathbf{h}_{test} \\
  &+ \Sigma^k_{i=1}\frac{W_V}{\sqrt{d}}\mathbf{h}_i\cdot (W_K\mathbf{h}_i)^{\mathrm{T}}\cdot \mathbf{h}_{test},
\end{aligned}
\end{equation}
\end{small}
where $\mathbf{h}_1,...,\mathbf{h}_k$ are the representations of demonstrations $X_1,...,X_k$. This formalization can be interpreted as introducing $k$ training samples $z_1=W_K\mathbf{h}_1,...,z_k=W_K\mathbf{h}_k$ to optimize the linear function $\mathcal{F}(z)$ simultaneously, where the gradient at each training sample $z_i$ satisfies
\begin{small}
\begin{equation}\label{meta_gradient_k}
\nabla_\mathcal{F} L(z_i,W_0)= \frac{W_V}{\sqrt{d}}\mathbf{h}_i.
\end{equation}
\end{small}

Similar to the proof of Theorem 1, we derive the following Theorem 2 to illustrate the condition of these samples to ensure a reduction in the loss of testing data $X_{test}$, where
\begin{small}
\begin{equation}\label{w_redefine}
\begin{aligned}
\hat{W}_{\epsilon,\bar{z}_k}\stackrel{def}{=}\operatorname*{argmin}_W \frac{1}{|\mathcal{D}_{pre}|}\Sigma_{z\in \mathcal{D}_{pre}}L(\mathcal{F}(z))+\epsilon\cdot \Sigma^k_{i=1}L(\mathcal{F}(z_i))
\end{aligned}
\end{equation}
\end{small}

\textbf{Theorem 2.} Assume $\nabla_\mathcal{F}L$ is Lipschitz continuous w.r.t $\mathcal{F}$ with constant $\mu$. If inequality~\eqref{condition_k} holds true, then $L(\mathbf{h}_{test},\hat{W}_{\frac{1}{|\mathcal{D}_{pre}|},\bar{z}_k})< L(\mathbf{h}_{test},\hat{W}_{0,\bar{z}_k})$, i.e., introducing training samples $\{z_1,...,z_k\}$ (i.e., demonstrations $X_1,...,X_k$) can reduce the testing loss on $\mathbf{h}_{test}$. 
\begin{small}
\begin{equation}\label{condition_k}
\begin{gathered}
\frac{k\lambda_{dd^\prime}}{\lambda_1}\|\nabla_W L(\mathbf{h}_{test},\hat{W})\|> \Sigma^k_{i=1}\|\mathbf{h}_{test}-z_i\|\cdot(\|\frac{W_V}{\sqrt{d}}\mathbf{h}_i\|+\mu C_1)
\end{gathered}
\end{equation}
\end{small}

Theorem 2 further indicates that the joint effect of different demonstrations follows an additive relationship. This implies that the selection of $k$ different demonstrations can be approximately considered independently.
\section{LMS3: Method Design}
\begin{figure*}[t]
	\centering
	\setlength{\abovecaptionskip}{1pt}
	\includegraphics[scale=0.37]{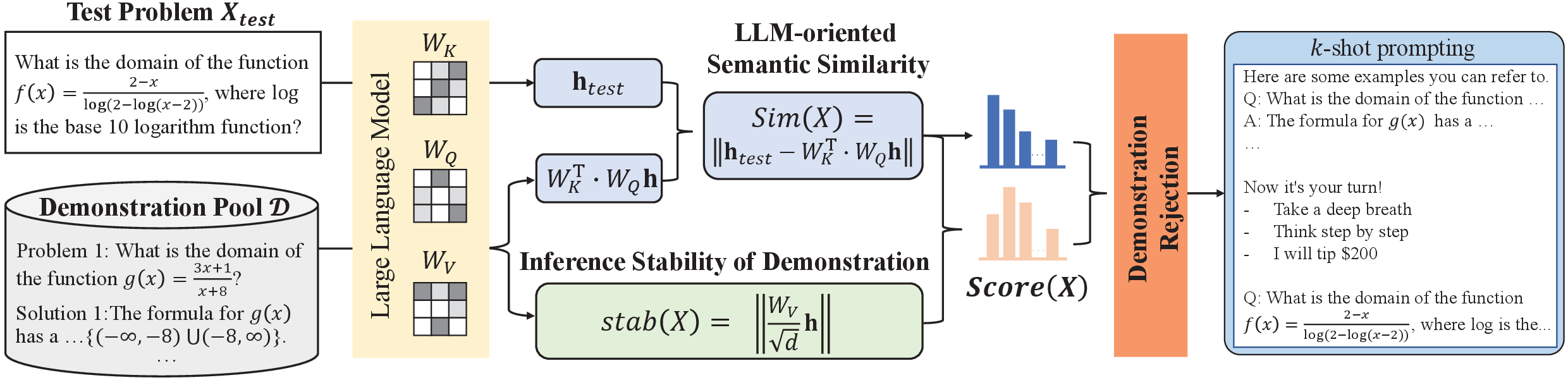}
	\caption{Illustration of our proposed LMS3 method.}
	\label{figure_method}
\vspace{-15pt}
\end{figure*}

Based on Section~\ref{section_theory}, an ideal demonstration $X$ needs to maximize the value of $L_1$ (i.e., minimize the empirical testing loss $L(\mathbf{h}_{test},\hat{W}_{\frac{1}{|\mathcal{D}_{pre}|},z_0})$ in Eq.~\eqref{taylor}). This is equivalent to minimize the right-hind side of Eq.~\eqref{condition} according to Eq.~\eqref{L1_final1} and can be further divided into: 1) minimize the value 
\begin{equation}\label{sim_score}
 Sim(X)\stackrel{def}{=}\|\mathbf{h}_{test}-W_K^{\mathrm{T}}\cdot W_Q\mathbf{h}\|,
\end{equation} 
(recall $z_0=W_K\mathbf{h}$ and $W_K$ is indeed $W_K^{\mathrm{T}}\cdot W_Q$ in the aforementioned section), and 2) minimize the value 
\begin{equation}\label{sta_score}
Stab(X)\stackrel{def}{=}\|\frac{W_V}{\sqrt{d}}\mathbf{h}\|.
\end{equation}. 

Specifically, $Sim(X)$ reflects a \textbf{LLM-oriented Semantic Similarity} between the demonstration $X$ and the test data $X_{test}$. It goes beyond traditional methods by taking into account 1) the whole reasoning path of demonstrations (recall $X$ includes both the problem context and the solution) and 2) the characteristics of the inference LLM itself, which is more consistent with intuition. The value of $Stab(X)$ is an evaluation of the \textbf{Inference Stability of Demonstration} $X$. Based on Eq~\eqref{meta_gradient}, $Stab(X)$ is indeed the length of gradient of the loss function on $X$. If $Stab(X)$ is low, it indicates that the LLM has reached a stable prediction with minimal loss on $X$, and the parameters will not be excessively altered due to the introduction of this sample. 

Since it is hard to simultaneously achieve the minimum of $Sim(X)$ and $Stab(X)$, two intuitive approximations are to minimize a demonstration scoring function that calculates their sum or product as follows:
\begin{equation}\label{main_score_sum}
 Score(X) = Sim(X) + \lambda_1 \cdot Stab(X),
\end{equation} 
\begin{equation}\label{main_score_mul}
 Score(X) = Sim(X) \cdot Stab(X),
\end{equation} 
However, Eq.~\eqref{main_score_sum} requires considering the scale differences between the two objectives and adjusting the hyperparameter $\lambda_1$ based on different LLMs and datasets, which is challenging to apply in practice. Therefore, we prefer Eq.~\eqref{main_score_mul} as the basic form of our scoring function. To implement $k$-shot in-context learning, we can select the top $k$ samples with the highest $Score(X)$ as demonstrations according to Theorem 2, which can ensure that the most relevant and stable samples are used to enhance the LLM's performance. 

Furthermore, we design a demonstration rejection mechanism, which is essential but has not yet been fully explored. For instance, it is possible that the examples with the highest $Score(X)$ still do not satisfy Eq.~\eqref{condition_k}. In such cases, unlike existing methods that always select top $k$ examples, we tend to refuse to provide any demonstration and instead use a zero-shot approach, because our theorems suggests that providing examples in this case will have a negative effect. We control $Sim(X)$ to achieve this rejection mechanism, because if an example's $Sim(X)$ is already too large, $Sim(X)\cdot \mu C_1$ might have exceeded the left-hand side of Eq.~\eqref{condition}. However, setting an absolute threshold for $Sim(X)$ is challenging since $\mu, C_1$ is unknown, and calculating the gradient norm $\|\nabla_W L(\mathbf{h}_{test},\hat{W})\|$ is costly. Therefore, we adopt a simplified relative threshold. We expect that the $Sim(X)$ of an ideal demonstration should be as small as possible relative to all examples. Consequently, we rank $Sim(X)$ of all candidate examples. If a demonstration $X$ ranked top-$k$ in $Score(X)$ does not have a $Sim(X)$ value within the top $\lambda$ smallest, we reject to select it.

\begin{table}[t]
\small
\centering
\setlength{\tabcolsep}{2.2pt}
\begin{tabular}{c|ccc@{\hskip -8pt}c}
\hline
\multirow{2}{*}{Method} & Theoretical & Dependency & \multirow{2}{*}{Generality} & \multirow{2}{*}{Complexity} \\
& Guarantee & on LLM & & \\
\hline 
Similar-ICL & \ding{55} & \ding{55} & \ding{51} & {\scriptsize $\mathcal{O}(\mathcal{M}+\mathcal{N})$}\\
Influence & \ding{55} & \ding{51} & \ding{51} & {\scriptsize $\mathcal{O}(\frac{\mathcal{MV}}{\mathcal{S}}+\mathcal{N})$} \\
InfICL & \ding{55} & \ding{51} & \ding{55} & {\scriptsize $\mathcal{O}(\mathcal{DM}+\mathcal{DV}+\mathcal{N})$} \\
TopK+MDL & \ding{55} & \ding{51} & \ding{55} & {\scriptsize $\mathcal{O}(\mathcal{M}+\mathcal{KN})$} \\
TopK+ConE & \ding{55} & \ding{51} & \ding{55} & {\scriptsize $\mathcal{O}(\mathcal{M}+\mathcal{KN})$} \\
IDS & \ding{55} & \ding{51} & \ding{51} & {\scriptsize $\mathcal{O}(\mathcal{M}+\mathcal{QN})$} \\
MI & \ding{55} & \ding{51} & \ding{55} & {\scriptsize $\mathcal{O}(\mathcal{MN})$} \\
SPELL & \ding{55} & \ding{51} & \ding{51} & {\scriptsize $\mathcal{O}(\mathcal{MR}+\mathcal{N})$} \\
\textbf{LMS3} (\textbf{ours}) & \ding{51} & \ding{51} & \ding{51} & {\scriptsize $\mathcal{O}(\mathcal{M}+\mathcal{N})$} \\
\hline
\end{tabular}
\setlength{\abovecaptionskip}{2pt}
\caption{Comparison of different methods, including Similar-ICL~\cite{liu2022makes,zhang2023retrieve,fu2022complexity,chen2024bge}, Influence~\cite{nguyen2023context} ($\mathcal{S}$ is the size of subset used to estimate influences, $\mathcal{V}$ is the size of validation set), InfICL~\cite{van2024context} ($\mathcal{D}$ is the number of parameters of external LLMs), TopK+MDL~\cite{wu2023self}, TopK+ConE~\cite{peng2024revisiting} ($\mathcal{K}$ is the number of candidate demonstrations), IDS~\cite{qin2023context} ($\mathcal{Q}$ is the number of iterations), MI~\cite{sorensen2022information}, SPELL~\cite{gonen2023demystifying} ($\mathcal{R}$ is the number of samples for estimating perplexity). The generality setting to \ding{55} indicates that these works are more suitable for classification tasks and hard to implement for mathematical reasoning task.}\label{comparison}
\vspace{-10pt}
\end{table}
\begin{table*}[ht]
\centering
\begin{tabular}{c|ccc|ccc}
\hline & \multicolumn{3}{c|} {Llama2-13B} & \multicolumn{3}{c} {Llama3-8B} \\
\cline { 2-4}\cline { 5-7}
& MAWPS & GSM8K & MATH & MAWPS & GSM8K & MATH \\
\hline 
zero-shot & 0.835{\scriptsize$\pm$ 0.009} & 0.414{\scriptsize$\pm$ 0.004} & 0.096{\scriptsize$\pm$ 0.005} & 0.951{\scriptsize$\pm$ 0.004} & 0.820{\scriptsize$\pm$ 0.016} & 0.324{\scriptsize$\pm$ 0.022}\\
\hline Random & 0.816{\scriptsize$\pm$ 0.004} & 0.405{\scriptsize$\pm$ 0.007} & 0.090{\scriptsize$\pm$ 0.010} & 0.951{\scriptsize$\pm$ 0.005} & 0.813{\scriptsize$\pm$ 0.003} & 0.330{\scriptsize$\pm$ 0.009}\\
Best-validate & 0.826{\scriptsize$\pm$ 0.001} & 0.410{\scriptsize$\pm$ 0.005} & 0.096{\scriptsize$\pm$ 0.007} & 0.932{\scriptsize$\pm$ 0.000} & 0.817{\scriptsize$\pm$ 0.008} & 0.332{\scriptsize$\pm$ 0.008}\\
\hline TF-IDF & 0.826{\scriptsize$\pm$ 0.021} & 0.424{\scriptsize$\pm$ 0.007} & 0.099{\scriptsize$\pm$ 0.006} & 0.945{\scriptsize$\pm$ 0.009} & 0.803{\scriptsize$\pm$ 0.007} & 0.344{\scriptsize$\pm$ 0.005}\\
BM25 & 0.815{\scriptsize$\pm$ 0.008} & 0.416{\scriptsize$\pm$ 0.014} & 0.098{\scriptsize$\pm$ 0.007} & 0.932{\scriptsize$\pm$ 0.003} & 0.805{\scriptsize$\pm$ 0.002} & 0.334{\scriptsize$\pm$ 0.004}\\
T5 & 0.810{\scriptsize$\pm$ 0.004} & \underline{0.426}{\scriptsize$\pm$ 0.013} & 0.093{\scriptsize$\pm$ 0.006} & 0.948{\scriptsize$\pm$ 0.021} & 0.817{\scriptsize$\pm$ 0.002} & 0.330{\scriptsize$\pm$ 0.009}\\
BGEM3 & 0.818{\scriptsize$\pm$ 0.013} & 0.407{\scriptsize$\pm$ 0.004} & 0.100{\scriptsize$\pm$ 0.011} & 0.938{\scriptsize$\pm$ 0.017} & 0.802{\scriptsize$\pm$ 0.000} & 0.340{\scriptsize$\pm$ 0.005}\\
OpenAI & 0.805{\scriptsize$\pm$ 0.014} & 0.416{\scriptsize$\pm$ 0.005} & 0.101{\scriptsize$\pm$ 0.002} & \underline{0.965}{\scriptsize$\pm$ 0.011} & 0.809{\scriptsize$\pm$ 0.008} & \underline{0.346}{\scriptsize$\pm$ 0.002}\\
\hline SPELL & 0.797{\scriptsize$\pm$ 0.009} & 0.394{\scriptsize$\pm$ 0.006} & 0.085{\scriptsize$\pm$ 0.003} & 0.945{\scriptsize$\pm$ 0.005} & \underline{0.821}{\scriptsize$\pm$ 0.008} & 0.343{\scriptsize$\pm$ 0.004}\\
Influence & 0.836{\scriptsize$\pm$ 0.010} & 0.405{\scriptsize$\pm$ 0.009} & \underline{0.102}{\scriptsize$\pm$ 0.000} & 0.929{\scriptsize$\pm$ 0.009} & 0.800{\scriptsize$\pm$ 0.015} & 0.333{\scriptsize$\pm$ 0.006}\\
IDS & \underline{0.839}{\scriptsize$\pm$ 0.005} & 0.424{\scriptsize$\pm$ 0.012} & 0.088{\scriptsize$\pm$ 0.001} & 0.920{\scriptsize$\pm$ 0.003} & 0.808{\scriptsize$\pm$ 0.001} & 0.330{\scriptsize$\pm$ 0.001}\\
\hline \textbf{LMS3} (\textbf{ours}) & \textbf{0.854}$^*${\scriptsize$\pm$ 0.008} & \textbf{0.447}$^*${\scriptsize$\pm$ 0.014} & \textbf{0.124}$^*${\scriptsize$\pm$ 0.003} & \textbf{0.966}{\scriptsize$\pm$ 0.010} & \textbf{0.837}$^*${\scriptsize$\pm$ 0.011} & \textbf{0.353}$^*${\scriptsize$\pm$ 0.002}\\
\hline
\end{tabular}
\caption{One-shot Answer Accuracy, with the best and runner-up methods highlighted in bold and underlined, respectively.}\label{performance_one_llama2}
\vspace{-18pt}
\end{table*}

Theoretically, to compute $Score(X)$, we need to input the concatenation of each ``(demonstration, testing data)'' pair $(X, X_{test})$ into the LLM to obtain their semantic representations $\mathbf{h},\mathbf{h}_{test}$. However, in practice, this process requires $\mathcal{O}(\mathcal{MN})$ complexity (measured by the number of LLM API calls) for testing, which significantly limits the efficiency. Therefore, we adopt an approximate by inputting each data individually into the LLM to obtain its semantic representation, reducing the testing complexity to $\mathcal{O}(\mathcal{M}+\mathcal{N})$ (the representations of all demonstrations can be pre-computed).

We illustrate the workflow of our method, named LMS3, in Figure~\ref{figure_method} and present the pseudo-code in \emph{Appendix 2}. LMS3 has several advantages as summarized in Table~\ref{comparison}. 1) \textbf{Theoretical Guarantee}: To the best of our knowledge, we are the first to theoretically quantify the impact of demonstrations on ICL reasoning performance and explain why and when they work. 2) \textbf{Rational Dependency}: Our analysis verifies that the optimal demonstration depends on the inference LLM (i.e., how the representations $\mathbf{h},\mathbf{h}_{test}$ are encoded). This is reasonable because a LLM's understanding of similar problems sets the upper limit on its ability to leverage these problems~\cite{peng2024revisiting}. Consequently, the optimal demonstration should be selected adaptively for different LLMs. However, existing methods like Similar-ICL estimate semantic similarity independently of the inference LLM and the demonstration is the same for all LLMs. 3) \textbf{Generalization Ability}: If we set $W_K^{\mathrm{T}}\cdot W_Q=I$ as an identity matrix and omit $Stab(X)$, our method degenerates into finding the demonstration with the closest semantic representation to the test data. This perspective unifies the current approaches, summarizing their main differences in the setting of $W_K^{\mathrm{T}}\cdot W_Q$ to obtain semantic representations. At the same time, our method, which analyzes the impact of demonstrations on the test loss, is not dependent on the task type. In addition to mathematical reasoning, it is also applicable to other generation tasks or classification tasks, which shows superior generalization ability. 4) \textbf{Low Complexity}: Compared to methods based on impact estimation or LLMs' feedback~\cite{van2024context,nguyen2023context,chang2023data}, our method does not require additional external LLMs, repeated testing of demonstration effects on validation set, or the computation of Hessian matrix, which brings much lower complexity. 

\section{Experiments}
\subsection{Experimental Setup}
\textbf{Datasets.} We use three datasets that cover a variety of types and difficulty levels. \textbf{MAWPS}~\cite{koncel2016mawps} consists of 2,373 elementary-level math word problems. \textbf{GSM8K}~\cite{cobbe2021training} is composed of 8,792 high-quality, more challenging elementary math problems with a higher number of steps. \textbf{MATH}~\cite{hendrycks2021measuring} is collected from high school math competition, containing 12,500 problems across seven categories such as algebra, geometry, and number theory, and is currently one of the most widely used benchmarks. Dataset partition and statistics are presented in \emph{Appendix 3}.

\noindent \textbf{Baselines.} We use Llama2-13B~\cite{touvron2023llama} and Llama3-8B~\cite{meta2024introducing} as the backbones to validate our method (please see \emph{Appendix} 4 for implementation details) and take 10 representative and SOTA baselines including:
\begin{itemize}
  \item \textbf{Random} randomly selects demonstrations from $\mathcal{D}$.
  \item \textbf{Best-validate} tests the performance of each data on a validation set, selecting the one with the highest accuracy,
\end{itemize}
and some typical Similar-ICL methods:
\begin{itemize}
  \item \textbf{TF-IDF} represents each problem as a TF-IDF vector, and selects the nearest sample to the test data.
  \item \textbf{BM25}~\cite{robertson2009probabilistic} selects demonstrations by retrieval method BM25.
  \item \textbf{T5}~\cite{raffel2020exploring} encodes problems with T5-large model and selects the most similar one.
  \item \textbf{BGEM3}~\cite{chen2024bge} integrate multiple information retrieval functionalities in a unified embedding.
  \item \textbf{OpenAI}~\cite{neelakantan2022text} adopts OpenAI Text-Embedding-3-Small model for problem representation,
\end{itemize}
as well as methods that do not rely on problem similarity:
\begin{itemize}
  \item \textbf{SPELL}~\cite{gonen2023demystifying} selects demonstrations by calculating their individual perplexity.
  \item \textbf{Influence}~\cite{nguyen2023context} divides $\mathcal{D}$ into multiple subsets. The preference of a demonstration is calculated by the difference in validation accuracy between subsets that include and exclude it. 
  \item \textbf{IDS}~\cite{qin2023context} iteratively selects training samples as demonstrations based on reasoning path similarity.
\end{itemize}
\subsection{Performance on One-shot Reasoning}
In Table~\ref{performance_one_llama2}, we present the performance of all methods in the one-shot setting. Firstly, it can be seen that our LMS3 outperforms all baselines across all datasets, and this effect is statistically significant with $p\leq 0.05$ (marked $*$). This directly indicates that the demonstrations chosen by our method better stimulate the LLM's contextual learning ability. Secondly, our LMS3 is the only one that consistently provides improvements over the zero-shot setting, while other methods exhibit certain fluctuations across different datasets. This can be attributed to our method being designed based on a theoretical analysis of when one-shot learning is effective (i.e., Theorem 1). These experimental results validate the rationality, effectiveness, and strong robustness of our theoretical findings. Thirdly, we observe that one-shot learning generally improves the backbone's performance on the more challenging MATH dataset, but sometimes shows a decrease on other datasets. We believe this is because the problems in MAWPS and GSM8K are relatively simple, and the LLM itself already has the capability to solve them. Introducing additional examples in this case might instead mislead the model's reasoning thought.
\subsection{Performance on Few-shot Reasoning}
\begin{figure}[t]
\centering
\setlength{\abovecaptionskip}{0pt}
\includegraphics[width=\linewidth]{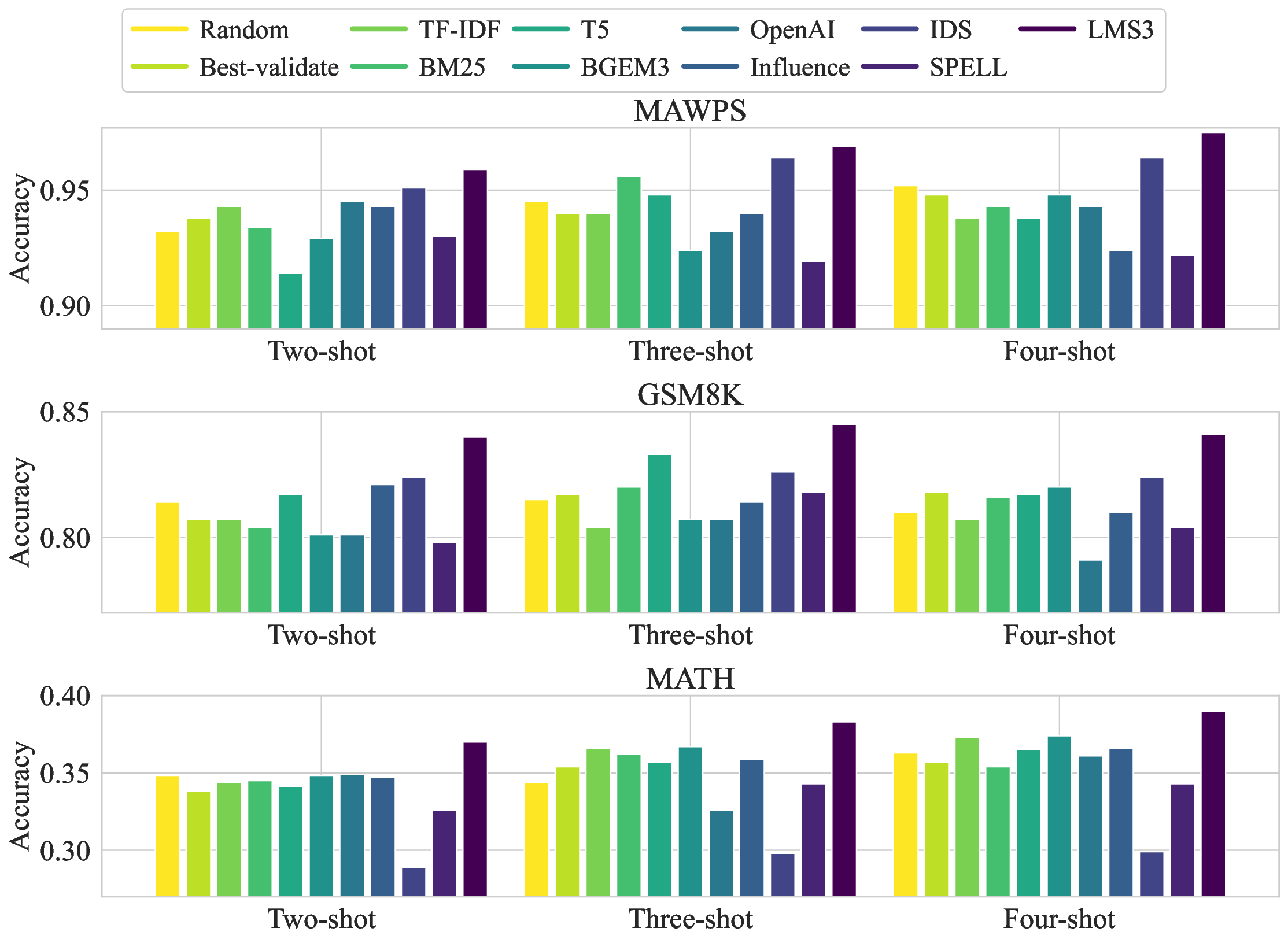}
\caption{Few-shot Answer Accuracy of Llama3-8B.}
\label{few_shot_exp}
\vspace{-12pt}
\end{figure}

Now we validate our LMS3 in the $k$-shot scenario, with Llama3-8B's performances at $k=\{2,3,4\}$ visualized in Figure~\ref{few_shot_exp}. Firstly, it indicates that our method remains superior across different settings, which is consistent with our Theorem 2, demonstrating the high applicability of our theorem to various scenarios. Secondly, as $k$ increases, the trend of reasoning performance varies across different datasets. Specifically, on MAWPS and MATH, the performances of most methods consistently improve with a higher $k$. However, on GSM8K, the accuracy for almost all methods declines after $k=3$. This highlights the need to balance the number and length of demonstrations, as an excessive number of demonstrations does not necessarily lead to increased accuracy. A dataset with longer problem lengths (i.e., GSM8K as indicated in \emph{Appendix} 3) may require fewer examples to achieve optimal performance.
\begin{figure}[t]
	\centering
	\setlength{\abovecaptionskip}{0pt}
	\includegraphics[scale=0.21]{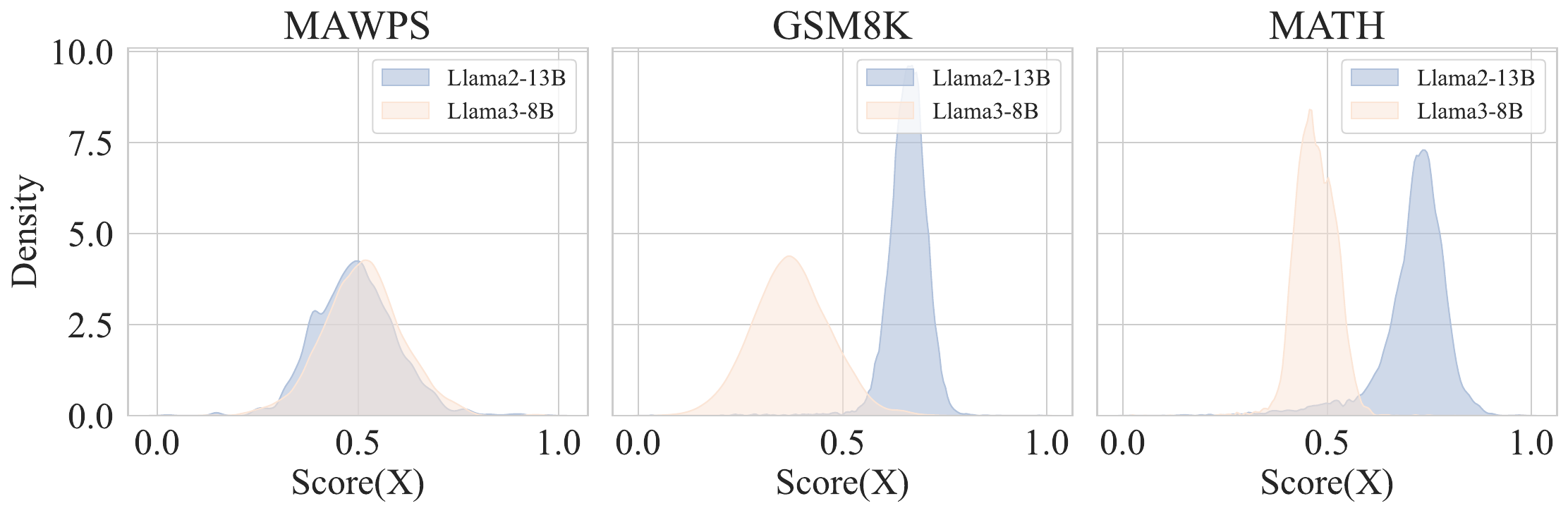}
	\caption{Distribution of $Score(X)$ in Eq.~\eqref{main_score_mul}.}
	\label{figure_scrore_llama3}
\vspace{-18pt}
\end{figure}

\subsection{LMS3 Analysis}
\noindent\textbf{Analysis of Scoring Function.} Figure~\ref{figure_scrore_llama3} presents the distribution of $Score(X)$ in Eq.~\eqref{main_score_mul} normalized by z-score, which verifies that our $Score(X)$ has good discriminative power for different samples. More importantly, we observe that the variances of the distributions for Llama2 on GSM8K and MATH, Llama3 on MATH, are relatively small. This indicates that the differences between samples in these cases are not significant, which can explain why most other one-shot baselines also perform better than the zero-shot setting in the corresponding columns of Table~\ref{performance_one_llama2}. In contrast, in other cases (e.g., on MAWPS), the performance gap between different samples is larger, and only our LMS3 can consistently achieve better results than zero-shot setting.
\begin{table}
\small
\centering
\setlength{\tabcolsep}{2pt} 
\resizebox{0.5\textwidth}{!}{
\begin{tabular}{c@{\hskip 0.002in}|ccc@{\hskip 0.002in}|ccc@{\hskip 0.020in}}
\hline
& \multicolumn{3}{c@{\hskip 0.002in}|}{ChatGPT} & \multicolumn{3}{c@{\hskip 0.002in}}{GPT-4} \\
\hline
& MAWPS & GSM8K & MATH & MAWPS & GSM8K & MATH \\
\hline 
zero-shot & 0.906 & 0.800 & 0.482 & 0.941 & 0.954 & \textbf{0.758}\\
Random & 0.858 & 0.839 & 0.503 & 0.976 & 0.946 & 0.702\\
Best-validate & 0.831 & 0.832 & \textbf{0.519} & 0.979 & 0.951 & 0.715\\
TF-IDF & 0.895 & 0.820 & 0.514 & 0.975 & 0.947 & 0.724\\
BM25 & 0.901 & 0.828 & 0.510 & \underline{0.987} & 0.953 & 0.691\\
T5 & 0.893 & 0.840 & 0.508 & 0.973 & 0.950 & 0.718\\
BGEM3 & 0.896 & 0.838 & 0.504 & 0.986 & 0.955 & 0.705\\
OpenAI & 0.898 & 0.829 & 0.513 & 0.979 & 0.945 & 0.699\\
Influence & 0.878 & \underline{0.848} & 0.515 & 0.974 & 0.955 & 0.702\\
IDS & \underline{0.908} & \underline{0.848} & 0.505 & 0.979 & \underline{0.959} & 0.742\\
\textbf{LMS3} (\textbf{ours}) & \textbf{0.909} & \textbf{0.862} & \underline{0.517} & \textbf{0.990} & \textbf{0.961} & \underline{0.752}\\
\hline
\end{tabular}
}
\setlength{\abovecaptionskip}{2pt}
\caption{Generalization performance on ChatGPT/GPT4.}\label{performance_chatgpt}
\vspace{-12pt}
\end{table}
\begin{figure}[t]
\centering
\setlength{\abovecaptionskip}{0pt}
\includegraphics[width=0.49\linewidth]{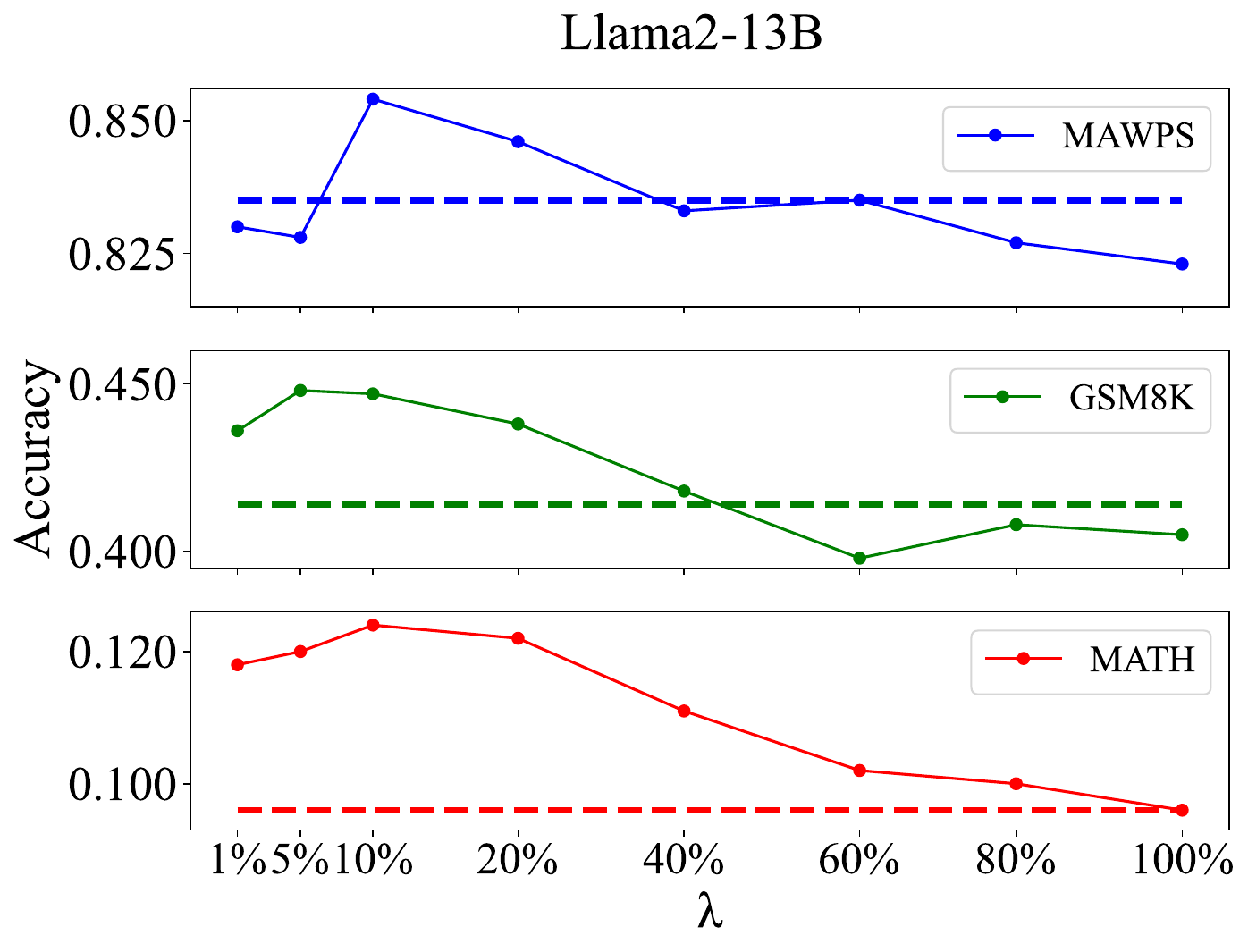}
\includegraphics[width=0.49\linewidth]{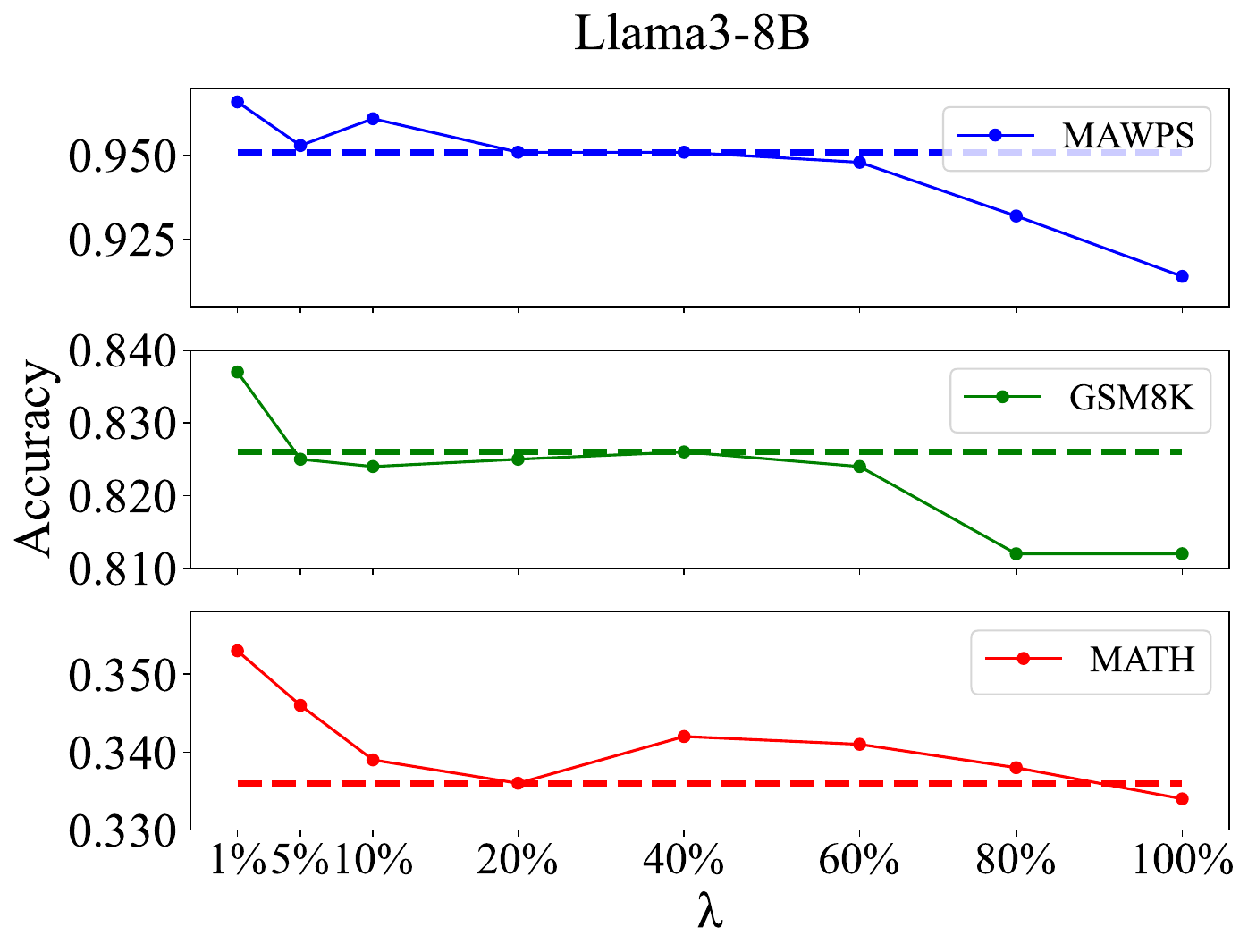}
\caption{Performance with varying $\lambda$. The dashed line corresponds to the result of the zero-shot setting.}
\label{hyper_exp}
\vspace{-15pt}
\end{figure}

Furthermore, to validate the necessity of our proposed demonstration rejection mechanism, we test the effects of $\lambda=\{1\%,5\%,10\%,20\%,40\%,60\%,80\%,100\%\}$. It is noteworthy that when $\lambda=100\%$, our rejection mechanism is essentially removed. From Figure~\ref{hyper_exp}, we can first observe that when $\lambda=100\%$, the accuracy of our LMS3 sometimes falls below that of the zero-shot results, which highlights the necessity of our rejection mechanism. Secondly, we notice that as $\lambda$ increases, the performance of Llama2 initially rises and then falls, while the performance of Llama3 consistently declines. On one hand, this indicates that the strength of $\lambda$ needs to be balanced differently for various LLMs, but this process is not complicated since the optimal $\lambda$ is basically within 10\%. On the other hand, this demonstrates that our $Sim(X)$ can effectively approximate the conditions in Theorems 1 and 2, as using it to implement the rejection mechanism can improve the model's performance.

\textbf{Generalization Ability.} One advantage of our method is that it reasonably considers the inference LLM when selecting demonstrations. However, it may raise concerns about its extensibility, as it requires access to the LLM's internal parameters. To address this issue, we offer the demonstrations selected by LMS3 (Llama3-8B) directly to the most advanced LLMs ChatGPT and GPT-4, compared with other methods that do not rely on the LLM's parameters. From Table~\ref{performance_chatgpt}, we can see that our LMS3 still achieves nearly optimal results, demonstrating the excellent generalization and representativeness of our selected demonstrations. Besides, we observe that all methods negatively impact the reasoning performance on MATH dataset when applied to GPT-4. We attribute this to that chain-of-thought examples may no longer be significantly beneficial for GPT-4, and future examples for GPT-4 might need to focus more on code~\cite{chen2023program,gao2023pal} or other formats.
\begin{figure}[t]
\centering
\setlength{\abovecaptionskip}{1pt}
\includegraphics[width=0.98\linewidth]{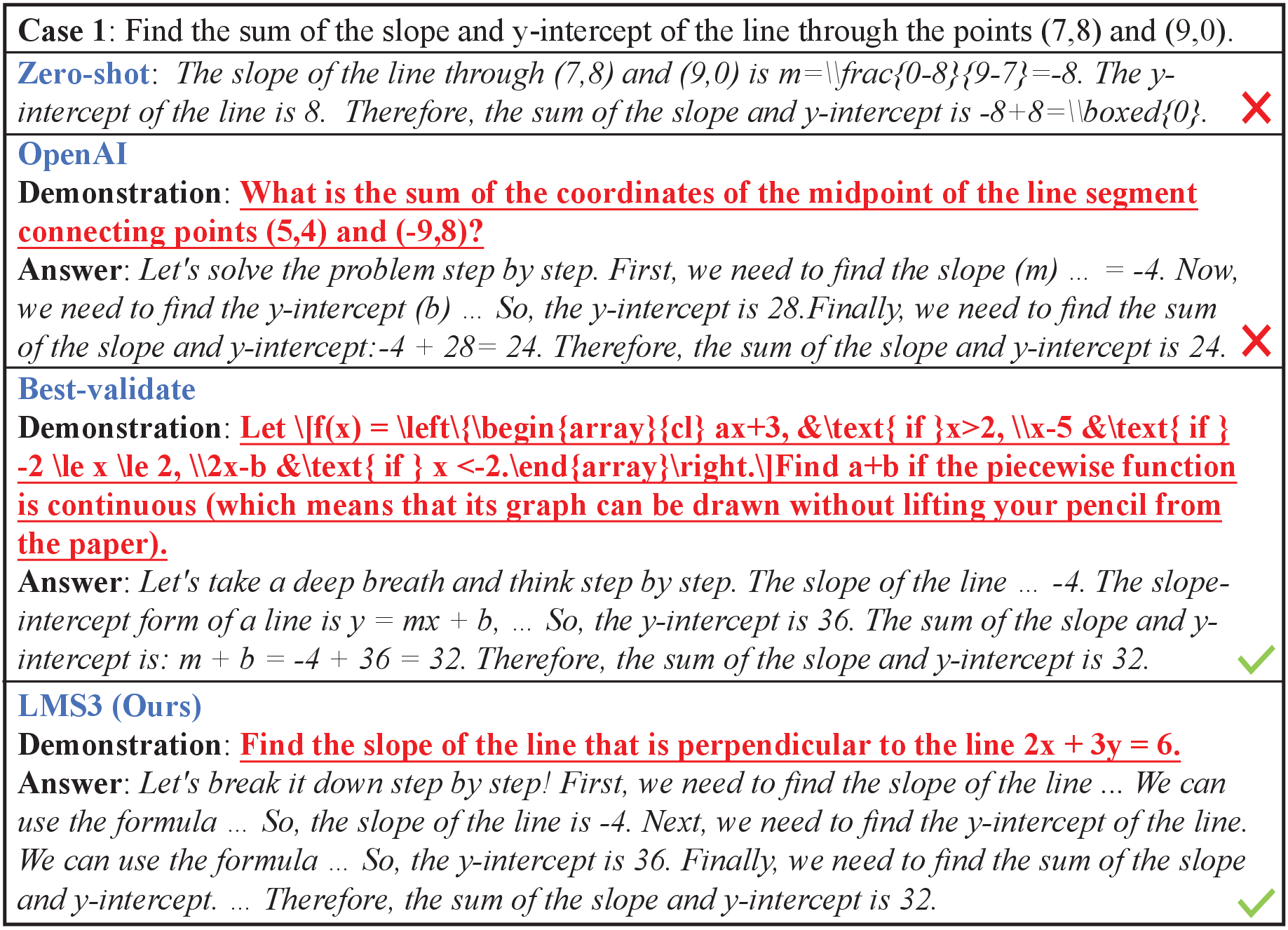}
\caption{Case Study (Case 1).}
\label{case1}
\vspace{-19pt}
\end{figure}

\textbf{Case Study.} We present three cases to validate the interpretability of our LMS3. Due to space limit, we show case 1 in Figure~\ref{case1} and cases 2 and 3 in \emph{Appendix 5}, and we omit the solutions in the demonstrations for brevity.

For cases 1 and 2, the baselines OpenAI and Best-validate both made errors. This indicates that considering only the semantic similarity of demonstrations or the effectiveness of demonstrations on the validation set alone, is insufficient. It is essential to balance similarity and the inference stability of demonstrations, as LMS3 does, to consistently achieve better results compared to zero-shot setting. In case 3, we again observe that the two baselines incorrectly answer a problem that zero-shot got right. In contrast, LMS3's rejection mechanism determines that the best demonstration still has a relatively large similarity distance $Sim(X)$ from the test sample (ranked in the top 1.19\% exceeding $\lambda=1\%$). By automatically rejecting this demonstration and adopting the zero-shot setting, LMS3 maintains the original performance, which verifies the necessity and effectiveness of our proposed demonstration rejection mechanism.
\section{Conclusion and Future Work} In this paper, we theoretically analyzed how demonstrations affected LLMs' mathematical reasoning performance. On this basis, we proposed a LMS3 method that balanced LLM-oriented semantic similarity and inference stability of demonstrations, and introduced a demonstration rejection mechanism to filter out negative situations. Experiments showed that our method was the only one to consistently improve the reasoning accuracy of LLMs, and our demonstrations exhibited strong generalization ability and interpretability. In the future, we will extend our method to more NLP tasks and apply our theory to broader scenarios. Please refer to \emph{Appendix 6} for more discussions and details.
\bibliography{bib}

\newpage
\section*{Paper Checklist}
This paper
\begin{itemize}
  \item Includes a conceptual outline and/or pseudocode description of AI methods introduced. (\textbf{Yes})
  \item Clearly delineates statements that are opinions, hypothesis, and speculation from objective facts and results. (\textbf{Yes})
  \item Provides well marked pedagogical references for less-familiare readers to gain background necessary to replicate the paper. (\textbf{Yes})
\end{itemize}

\noindent Does this paper make theoretical contributions? (\textbf{Yes})

\noindent If yes, please complete the list below.
\begin{itemize}
  \item All assumptions and restrictions are stated clearly and formally. (\textbf{Yes})
  \item All novel claims are stated formally (e.g., in theorem statements). (\textbf{Yes})
  \item Proofs of all novel claims are included. (\textbf{Yes})
  \item Proof sketches or intuitions are given for complex and/or novel results. (\textbf{Yes})
  \item Appropriate citations to theoretical tools used are given. (\textbf{Yes})
  \item All theoretical claims are demonstrated empirically to hold. (\textbf{Yes})
  \item All experimental code used to eliminate or disprove claims is included. (\textbf{Yes})
\end{itemize}

\noindent Does this paper rely on one or more datasets? (\textbf{Yes})

\noindent If yes, please complete the list below.
\begin{itemize}
  \item A motivation is given for why the experiments are conducted on the selected datasets. (\textbf{Yes})
  \item All novel datasets introduced in this paper are included in a data appendix. (\textbf{Yes})
  \item All novel datasets introduced in this paper will be made publicly available upon publication of the paper with a license that allows free usage for research purposes. (\textbf{Yes})
  \item All datasets drawn from the existing literature (potentially including authors' own previously published work) are accompanied by appropriate citations. (\textbf{Yes})
  \item All datasets drawn from the existing literature (potentially including authors' own previously published work) are publicly available. (\textbf{Yes})
  \item All datasets that are not publicly available are described in detail, with explanation why publicly available alternatives are not scientifically satisficing. (\textbf{NA})
\end{itemize}

\noindent Does this paper include computational experiments? (\textbf{Yes})

\noindent If yes, please complete the list below.
\begin{itemize}
  \item Any code required for pre-processing data is included in the appendix. (\textbf{Yes})
  \item All source code required for conducting and analyzing the experiments is included in a code appendix. (\textbf{Yes})
  \item All source code required for conducting and analyzing the experiments will be made publicly available upon publication of the paper with a license that allows free usage for research purposes. (\textbf{Yes})
  \item All source code implementing new methods have comments detailing the implementation, with references to the paper where each step comes from. (\textbf{Yes})
  \item If an algorithm depends on randomness, then the method used for setting seeds is described in a way sufficient to allow replication of results. (\textbf{Yes})
  \item This paper specifies the computing infrastructure used for running experiments (hardware and software), including GPU/CPU models; amount of memory; operating system; names and versions of relevant software libraries and frameworks. (\textbf{Yes})
  \item This paper formally describes evaluation metrics used and explains the motivation for choosing these metrics. (\textbf{Yes})
  \item This paper states the number of algorithm runs used to compute each reported result. (\textbf{Yes})
  \item Analysis of experiments goes beyond single-dimensional summaries of performance (e.g., average; median) to include measures of variation, confidence, or other distributional information. (\textbf{Yes})
  \item The significance of any improvement or decrease in performance is judged using appropriate statistical tests (e.g., Wilcoxon signed-rank). (\textbf{Yes})
  \item This paper lists all final (hyper-)parameters used for each model/algorithm in the paper's experiments. (\textbf{Yes})
  \item This paper states the number and range of values tried per (hyper-) parameter during development of the paper, along with the criterion used for selecting the final parameter setting. (\textbf{Yes})
\end{itemize}
\end{document}